\documentclass[sigconf]{acmart}
\AtBeginDocument{%
  \providecommand\BibTeX{{%
    \normalfont B\kern-0.5em{\scshape i\kern-0.25em b}\kern-0.8em\TeX}}}

\usepackage{wrapfig}
\usepackage{multirow}
\usepackage{makecell}
\usepackage{tikz}
\usepackage{enumitem}

\newcommand{\challenge}{heterogeneity}
\newcommand{\challengeit}{\challenge}
\newcommand{\hh}{hardware heterogeneity}
\newcommand{\bh}{state heterogeneity}

\newcommand{\Hh}{Hardware heterogeneity}
\newcommand{\Bh}{State heterogeneity}

\newcommand{\ha}{heterogeneity-aware}
\newcommand{\hua}{heterogeneity-unaware}
\newcommand{\dataset}{M-Type}

\newcommand{\reviewer}[1]{
}

\definecolor{airforceblue}{rgb}{0.36, 0.54, 0.66}

\newcommand*\blackcircled[1]{\tikz[baseline=(char.base)]{
		\node[shape=circle,fill=black,draw,text=white,inner sep=1pt] (char) {#1};}}

\newcommand{\bfsubsubsection}[1]{\subsubsection{\textbf{#1}}}

\copyrightyear{2021}
\acmYear{2021}
\setcopyright{iw3c2w3}
\acmConference[WWW '21]{Proceedings of the Web Conference 2021}{April 19--23, 2021}{Ljubljana, Slovenia}
\acmBooktitle{Proceedings of the Web Conference 2021 (WWW '21), April 19--23, 2021,
Ljubljana, Slovenia}
\acmPrice{}
\acmDOI{10.1145/3442381.3449851}
\acmISBN{978-1-4503-8312-7/21/04}

\begin{document}
\fancyhead{}

\title{Characterizing Impacts of Heterogeneity in Federated Learning upon Large-Scale Smartphone Data}
\renewcommand{\shorttitle}{Characterizing Impacts of Heterogeneity in Federated Learning}


\author{Chengxu Yang, Qipeng Wang}
\affiliation{%
    \institution{Key Lab of High Confidence Software Technologies (Peking University), MoE}
    \country{Beijing, China}
}
\email{yangchengxu@pku.edu.cn} \email{wangqipeng@stu.pku.edu.cn}


\author{Mengwei Xu}
\affiliation{%
    \institution{State Key Laboratory of Networking and Switching Technology (BUPT)}
  \institution{Key Lab of High Confidence Software Technologies (Peking University), MoE}
  \country{Beijing, China}
}
\email{mwx@bupt.edu.cn}

\author{Zhenpeng Chen}
\affiliation{%
  \institution{Key Lab of High Confidence Software Technologies (Peking University), MoE}
  \country{Beijing, China}
}
\email{czp@pku.edu.cn}

\author{Kaigui Bian}
\affiliation{%
  \institution{Department of Computer Science and Technology, Peking University}
  \country{Beijing, China}
}
\email{bkg@pku.edu.cn}

\author{Yunxin Liu}
\affiliation{%
  \institution{Microsoft Research}
  \country{Beijing, China}
}
\email{yunxin.liu@microsoft.com}

\author{Xuanzhe Liu}\authornote{Corresponding author: Xuanzhe Liu (xzl@pku.edu.cn).}
\affiliation{%
  \institution{Key Lab of High Confidence Software Technologies (Peking University), MoE}
  \country{Beijing, China}
}
\email{xzl@pku.edu.cn}


\begin{abstract}
\noindent Federated learning (FL) is an emerging, privacy-preserving machine learning paradigm, drawing tremendous attention in both academia and industry.
A unique characteristic of FL is \textit{heterogeneity}, which resides in the various hardware specifications and dynamic states across the participating devices.
Theoretically, heterogeneity can exert a huge influence on the FL training process, e.g., causing a device unavailable for training or unable to upload its model updates.
Unfortunately, these impacts have never been systematically studied and quantified in existing FL literature.

In this paper, we carry out the first empirical study to characterize the impacts of heterogeneity in FL.
We collect large-scale data from 136k smartphones that can faithfully reflect heterogeneity in real-world settings.
We also build a \textit{heterogeneity-aware} FL platform that complies with the standard FL protocol but with heterogeneity in consideration.
Based on the data and the platform, we conduct extensive experiments to compare the performance of state-of-the-art FL algorithms under \textit{heterogeneity-aware} and \textit{heterogeneity-unaware} settings.
Results show that heterogeneity causes non-trivial performance degradation in FL, including up to 9.2\% accuracy drop, 2.32$\times$ lengthened training time, and undermined fairness.
Furthermore, we analyze potential impact factors and find that \textit{device failure} and \textit{participant bias} are two potential factors for performance degradation.
Our study provides insightful implications for FL practitioners.
On the one hand, our findings suggest that FL algorithm designers consider necessary heterogeneity during the evaluation.
On the other hand, our findings urge system providers to design specific mechanisms to mitigate the impacts of heterogeneity.

\end{abstract}

\ccsdesc[500]{Human-centered computing~Ubiquitous and mobile computing}
\ccsdesc[500]{Computing methodologies~Artificial intelligence}
\ccsdesc[500]{Security and privacy~Privacy protections}

\keywords{Federated learning; heterogeneity; measurement study}


\maketitle

\section{Introduction}\label{sec:intro}

In the past few years, we have witnessed the increase of machine learning (ML) applications deployed on mobile devices~\cite{xu2019first,CHENDLDEPLOY,CHENmobile,deepcache,deeptype}. These applications usually need to collect personal user data to train ML models. However, due to the increasing concerns of user privacy, e.g., the recent released GDPR~\cite{GDPR} and CCPA~\cite{CCPA},
personal data cannot be arbitrarily collected and used without permission granted~\cite{wwwMeurischBM20}. 
Therefore, various privacy-preserving ML techniques have been proposed~\cite{wwwArcherLLV17,mcmahan2016communication,wwwChenLALML18}, where the emerging federated learning (FL) has drawn tremendous attentions~\cite{li2018federated,li2019federated}.
The key idea of FL is to employ a set of personal mobile devices, e.g., smartphones, to train an ML model collaboratively under the orchestration of a central parameter server. Since the training process of FL takes place on mobile devices (i.e., on-device training) without uploading user personal data outside devices, it is considered to be quite promising for preserving user privacy in ML applications. 


Recently, various emerging FL algorithms have been proposed, e.g., \textit{FedAvg} \cite{mcmahan2016communication}, \textit{Structured Updates}~\cite{konevcny2016federatedstrategies}, and \textit{q-FedAvg} \cite{li2019fair}. 
To evaluate these algorithms, existing FL studies typically take a simulation approach~\cite{mcmahan2016communication, chen2019communication,li2019fair,bagdasaryan2018backdoor} given the high cost of real deployment.
However, they have no data to describe how devices participate in FL\footnote{Although Google has built a practical FL system ~\cite{bonawitz2019towards}, its detailed data are not disclosed.}.
As a result, they usually have an overly ideal assumption, i.e., all the devices are always available for training and equipped with \textit{homogeneous} hardware specifications (e.g., the same CPU and RAM capacity)~\cite{mcmahan2016communication, konevcny2016federatedstrategies, li2019fair,bagdasaryan2018backdoor,mohri2019agnostic,jiang2019improving}.

However, these assumptions could be too ideal for FL deployment in practice.
More specifically, FL usually requires a substantial number of devices to collaboratively accomplish a learning task, which poses a unique challenge, namely \textbf{\challengeit{}} \cite{li2019federated}.
In practice, the heterogeneity can be attributed to two major aspects:
(1) One is from hardware specifications of devices (called \textit{\hh{}}), e.g., different CPU, RAM, and battery life.
(2) Additionally, the state and running environment of participating devices can be various and dynamic (called \textit{\bh{}}), e.g., CPU busy/free, stable/unreliable network connections to the server, etc.

Intuitively, heterogeneity can impact FL in terms of accuracy and training time. For instance, it is not surprising when a device fails to upload its local model updates to the server (called \textit{device failure}), which can definitely affect the training time to obtain a converged global model. 
Furthermore, devices that seldom participate in an FL task due to abnormal states, e.g., CPU busy, can be underrepresented by the global model.

Although some recent studies~\cite{li2018federated, nishio2019client, laguel2020device, chai2019towards} have realized the heterogeneity in FL, its impacts have never been comprehensively quantified over large-scale real-world data.
In this paper, we present the first empirical study to demystify the impacts of \challengeit{} in FL tasks.
To this end, we develop a holistic platform that complies with the standard and widely-adopted FL protocol ~\cite{bonawitz2019towards, yang2018applied, hard2018federated}, but for the first time, facilitates reproducing existing FL algorithms under \textit{\ha} settings, i.e., devices have dynamic states and various hardware capacities.
Undoubtedly, conducting such a study requires the data that can faithfully reflect the \challengeit{} in real-world settings. Therefore, we collect the device hardware specifications and regular state changes (including the states related to device check-in and drop-out) of 136k smartphones in one week through a commodity input method app (IMA).
We then plug the data into our \textit{\ha{}} platform to simulate the device state dynamics and hardware capacity.

Based on the data and platform, we conduct extensive measurement experiments to compare the state-of-the-art FL algorithms' performance, including model accuracy and training time under \ha{} and \hua~settings.
We select four typical FL tasks: two image classification tasks and two natural language processing tasks. For every single task, we employ a benchmark dataset for model training. Three of the benchmark datasets~\cite{reddit,celeba,cohen2017emnist} have been widely used in existing FL-related studies~\cite{caldas2018leaf,li2019fair,li2018federated,konevcny2016federatedstrategies,mcmahan2016communication}, and the last one is a real-world text input dataset collected from the aforementioned IMA.

\noindent $\bullet$ \textbf{Findings.} Heterogeneity leads to non-trivial impacts on the performance of FL algorithms, including accuracy drop, increased training time, and undermined fairness. For the basic algorithm ($\S\ref{subsec:basic}$), i.e., \textit{FedAvg}~\cite{mcmahan2016communication}, when \challengeit{} is considered, its performance is compromised in terms of 3.1\% accuracy drop (up to 9.2\%) and 1.74$\times$ training time (up to 2.32$\times$) on average. For other advanced algorithms ($\S\ref{subsec:advance}$), i.e., gradient compression algorithms, including \textit{Structured Updates}~\cite{konevcny2016federatedstrategies}, \textit{Gradient Dropping}~\cite{GDrop}, and \textit{SignSGD}~\cite{SignSGD}, and advanced aggregation algorithms, i.e., \textit{q-FedAvg}~\cite{li2019fair} and \textit{FedProx}~\cite{li2018federated}, optimizations are not always effective as reported. For example, \challengeit{} hinders \textit{q-FedAvg} from addressing the fairness issues in FL.
We also find that current gradient compression algorithms can hardly speed up FL convergence under \ha{} settings.
In the worst case, the training time is lengthened by 3.5$\times$. These findings indicate that \challengeit{} cannot be simply ignored when designing FL algorithms.

\noindent $\bullet$ \textbf{Analysis of Potential Impact Factors}. 
We first break down the \challengeit{} to analyze the individual impacts of \bh{} and \hh, respectively ($\S\ref{subsec:breakdown}$).
We find that both types of \challengeit{} can slow down the learning process, while \bh{} is often more responsible for the accuracy degradation.
Then we zoom into our experiments and find out two major factors that are particularly obvious under \ha{} settings.
(1) \textit{Device failure} ($\S\ref{subsec:failure}$): On average, 11.6\% of selected devices fail to upload their model updates per round due to unreliable network, excessive training time, and drop-out caused by user interruption.
This failure slows down the model convergence and wastes valuable hardware resources.
(2) \textit{Participant bias} ($\S\ref{subsec:bias}$): Devices attend FL process in a biased manner. For instance, we find that more than 30\% of devices never participate in the learning process when the model converges and the global model is dominated by active devices (top 30\% devices contribute to 81\% total computation). \Bh{} is the major cause for the participant bias.

Our extensive experiments provide several insightful implications as summarized in $\S$\ref{sec:impli}.
For instance, {FL algorithm designers} should consider necessary \challengeit{} in the evaluation environment of FL, while \textit{FL system providers} should design specific mechanisms to mitigate the impacts of \challengeit{}.
In summary, the major contributions of this paper are as follows:

\begin{itemize}[leftmargin=*]
\item We build a \ha{} FL platform with a large-scale dataset collected from 136k smartphones, which can help simulate the state and hardware heterogeneity for exploring FL in real-world practice\footnote{We have released our dataset and source code at  \url{https://github.com/PKU-Chengxu/FLASH} to facilitate future FL research.}.
\item We conduct extensive measurement experiments to demystify the non-trivial impacts of \challengeit{} in existing FL algorithms.
\item We make an in-depth analysis of possible factors for impacts introduced by heterogeneity. Our results can provide insightful and actionable implications for the research community.
\end{itemize}

\begin{figure*}
    \centering
    \includegraphics[width=0.95\linewidth]{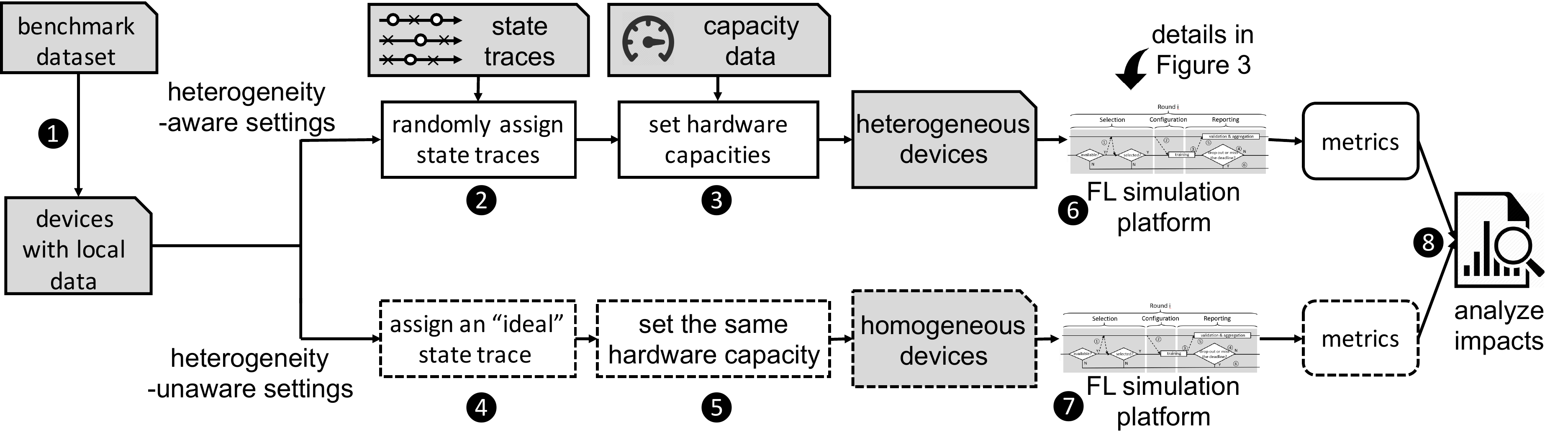}
    \vspace{-5pt}
    \caption{Overview of our methodology.}
    \label{fig:workflow}
    \vspace{-10pt}
\end{figure*}

\section{Background and related work}\label{sec:background}
\noindent \textbf{FL} is an emerging privacy-preserving learning paradigm.
Among different FL scenarios~\cite{kairouz2019advances}, we focus on a widely studied one, i.e., \textit{cross-device} FL, which utilizes a federation of \textit{client devices}\footnote{In the rest of this paper, we use \textit{device} to refer to \textit{client device}.}, coordinated by a \textit{central server}, to train a global ML model.
A typical FL workflow \cite{bonawitz2019towards} consists of many rounds, where each round can be divided into three phases: (1) the central server first selects devices to participate in the FL; (2) each selected device retrieves the latest global model from the server as the local model, re-trains the local model with local data, and uploads the updated weights/gradients of the local model to the server; (3) the server finally aggregates the updates from devices and obtains a new global model. 

In practice, FL is typically implemented based on state-of-the-art FL algorithms, such as \textit{FedAvg}~\cite{mcmahan2016communication}. 
In \textit{FedAvg}, devices perform multiple local training epochs, where per round a device updates the weights of its local model using its local data.
Then the central server averages the updated weights of local models as the new weights of the global model.
\textit{FedAvg} is a representative FL algorithm that has been widely used in existing FL-related studies~\cite{li2019fair, mohri2019agnostic, jiang2019improving, konevcny2016federatedstrategies} and also deployed in the industry, e.g., in Google's production FL system~\cite{chai2019towards}. 
Therefore, we use \textit{FedAvg} as the basic FL algorithm to study \challengeit{}'s impacts ($\S\ref{subsec:basic}$).

In addition, many advanced algorithms have been proposed to optimize FL, including reducing the communication cost between the central server and devices~\cite{konevcny2016federatedstrategies,smith2017federated,chen2019communication,reisizadeh2019fedpaq,bonawitz2019federated,yuan2020hierarchical}, enhancing privacy guarantee~\cite{bonawitz2017practical,mcmahan2017learning,bagdasaryan2018backdoor,melis2019exploiting, nasr2018comprehensive},
ensuring fairness across devices~\cite{li2019fair,mohri2019agnostic, jiang2019improving}, minimizing on-device energy cost~\cite{lismartpc}, etc. However, most of them have not been well evaluated in a \ha{} environment, making their benefits unclear in real-world deployment. 
Therefore, we make the first attempt to study how \challengeit{} impacts the effectiveness of these advanced FL algorithms ($\S\ref{subsec:advance}$).



\textbf{Heterogeneity} is considered as one of the core challenges in FL~\cite{li2019federated}.
Some existing work~\cite{li2018federated, nishio2019client, laguel2020device, chai2019towards} has studied the heterogeneity in FL but not in a comprehensive way.
In particular, they ignore \bh{} and randomly set training time to simulate \hh{}, leaving communication capacities unconsidered.
For example, \textit{FedProx}~\cite{li2018federated} handles the \hh{} by allowing each participating device to perform a variable amount of work, but the hardware capability of each device is randomly set and state changes of devices remain unconsidered.
\textit{FedCS}~\cite{nishio2019client} accelerates FL by managing devices based on their resource conditions and allowing the server to aggregate as many device updates as possible, but it assumes that the network is stable and not congested, and randomly sets the training time from 5 to 500 seconds.
Chai et al.~\cite{chai2019towards} have studied the impacts of \hh{} by allocating varied CPU resources when simulating FL, but they leave \bh{} unconsidered.
Our study differs from existing work in two aspects: (1) we comprehensively consider \hh{} and \bh{} in the experimental environment powered by real-world data, and (2) we build an FL scenario with a much larger device population (up to 136k).


\section{The Measurement Approach}
\label{sec:method}

\subsection{Approach Overview}
\label{subsec:workflow}
\noindent Figure~\ref{fig:workflow} illustrates the overall workflow of our measurement approach.
It starts from a \textit{benchmark dataset} that is typically partitioned into thousands or millions of devices holding their own local data for training (\blackcircled{1}).
For a fair comparison, we always use the same partition strategy in the \ha{} settings and \hua{} settings, i.e., the local training data on a given device are the same.

For \ha{} settings, we randomly assign a state trace (\blackcircled{2}) and a hardware capacity (\blackcircled{3}) to each device.
A state trace determines whether a device is available for local training at any simulation timestamp, while the hardware capacity specifies the training speed and communication bandwidth.
Both datasets are collected from large-scale real-world mobile devices through an IMA app (details in $\S$\ref{dataset}).
As a result, we get a heterogeneous device set with different local training data, hardware capacities, and state change dynamics.

For \hua{} settings, we assign each device with an ``ideal'' state trace, i.e., the device always stays available for local training and never drops out (\blackcircled{4}), and a uniform hardware capacity as the mid-end device in our IMA dataset (Redmi Note 8) (\blackcircled{5}).
As a result, we get a homogeneous device set with the same hardware capacity and state change dynamics, as existing FL platforms do.

We next deploy the two device sets to our FL simulation platform and execute the FL task (e.g., image classification) under the same configurations (\blackcircled{6} and \blackcircled{7}).
The simulation platform extends the standard FL protocol with heterogeneity consideration, e.g., a device can quit training due to a state change (details in $\S$\ref{subsec:fl-runtime}).
We finally analyze \challengeit{}'s impacts by comparing the metric values achieved by heterogeneous devices and homogeneous devices (\blackcircled{8}).

\subsection{The Datasets}\label{dataset}
\noindent As described in $\S\ref{subsec:workflow}$, we use two types of datasets in this study, including (1) the \textit{IMA dataset} describing the \challengeit{} in real-world smartphone usage, and (2) benchmark datasets containing devices' local data used for training and testing ML models.

\bfsubsubsection{IMA dataset}
\label{subsec:datasets}
To power the heterogeneity-aware settings, we collect large-scale real-world data from a popular IMA that can be downloaded from Google Play.
The dataset can be divided into two parts, including (1) \textit{device state traces} for annotating state heterogeneity, and (2) \textit{capacity data} for annotating hardware heterogeneity.

\noindent $\bullet$ \textbf{Device state traces} record the state changes (including battery charging, battery level, network environment, screen locking, etc.) of 136k devices within one week starting from Jan. 31 in 2020.
More specifically, every time the aforementioned state changes, the IMA records it with the timestamp and saves it as a state entry (refer to Table~\ref{tab:trace}).

In total, we collect 136k traces (one for each device) containing 180 million state entries, accounting for 111GB of storage.

\begin{figure}
    \centering
    \includegraphics[width=\linewidth]{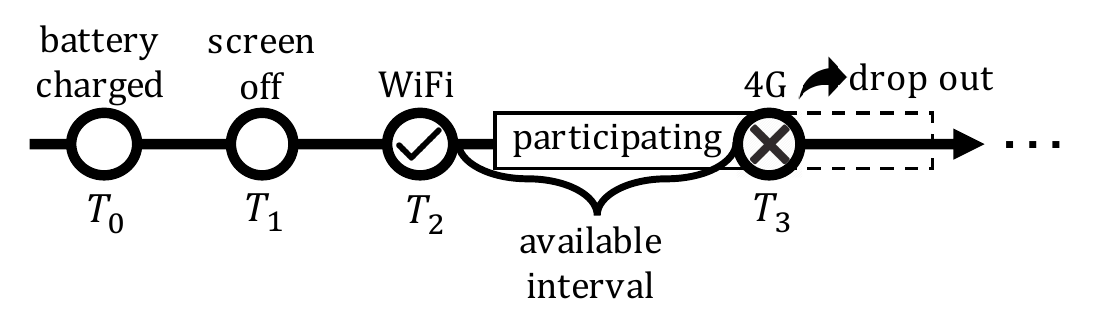}
    \caption{A trace is a series of state changes over time.}
    \label{fig:trace_usage}
    \vspace{-15pt}
\end{figure}

The state traces are to determine the time intervals when a device is available for local training, which are critical in understanding the FL performance under \ha{} settings.
Figure~\ref{fig:trace_usage} concretely exemplifies how a trace works during the simulation.
The device becomes available for training at $T_2$ because it meets the state criteria~\cite{bonawitz2019towards}, i.e., when a device is idle, charged, and connected to WiFi.
Then after a period of time at $T_3$, the network environment changes to ``4G'', thus the device becomes unavailable.
As a result, we obtain a training-available interval between $T_2$ and $T_3$.


As far as we know, this is the first-of-its-kind device usage dataset collected from large-scale real-world devices, making it much more representative than the datasets covering a small group of devices~\cite{20users,yogesh2014short}.

\begin{table}[]
    \centering
    \scalebox{0.75}{
    \begin{tabular}{|l|l|l|}
        \toprule
         Field & Description & Example\\
         \hline
         user\_id & Anonymized user id. & xxxyyyzzz\\
         device\_model & device type & SM-A300M\\
         screen\_trace & screen on or off & screen\_on\\
         screen\_lock\_trace & screen lock or unlock & screen\_lock\\
         time & time at current state & 2020-01-29 05:52:16\\
         network\_trace & network condition & 2G/3G/4G/5G/WiFi\\
         battery\_trace & battery charging state, battery level & battery\_charged\_off 96.0\%\\
         \bottomrule
    \end{tabular}
    }
    \caption{Example of a state entry.}
    \label{tab:trace}
    \vspace{-10pt}
\end{table}

\noindent $\bullet$ \textbf{Hardware capacity data} indicate the computational and communication capacities of different devices.
This dataset, along with the aforementioned state trace, determines how long and whether a device can successfully finish its local training and upload the model updates to the central server before a deadline.

For the computational capacity, we seek to obtain the training speed for each given device.
However, our collected IMA dataset contains more than one thousand types of devices, making it rather difficult to profile.
Thus, we employ a ``clustering'' approach by mapping all device models to a small number of representative device models that we afford to offline profile.
The mapping consists of two steps:
(1) The total device models are first mapped to the device models profiled by AI-Benchmark~\cite{ignatov2018ai}, a comprehensive AI performance benchmark.
For a few device models that AI-Benchmark does not cover, we make a random mapping.
It reduces the number of device models to 296.
(2) The remaining device models are then mapped to what we afford to profile.
So far, we have profiled three representative and widely-used device models (Samsung Note 10, Redmi Note 8, and Nexus 6), and we plan to include more device models in the future.
To profile these devices, we run on-device training using the open-source ML library DL4J \cite{dl4j} and record their training time for each ML model used in our experiments.
We are aware of learning-based approaches~\cite{deepwear} to obtain the on-device ML performance, but our empirical efforts show that these approaches are not precise enough for on-device training tasks.

For the communication capacity, we recruit 30 volunteers and deploy a testing app on their devices to periodically obtain (i.e., every two hours) the downstream/upstream bandwidth between the devices and a cloud server.
We fit each volunteer's data to a normal distribution and randomly assign a distribution to the device during the simulation.
The bandwidth data determine the model uploading/downloading time during simulation.

\bfsubsubsection{Benchmark datasets}
\label{subsec:benchmark}
We use four benchmark datasets to quantitatively study the impacts of heterogeneity on FL performance.
Three of them (i.e., Reddit~\cite{reddit}, Femnist~\cite{cohen2017emnist}, and Celeba~\cite{celeba}) are synthetic datasets widely adopted in FL literature~\cite{li2019fair,li2019federated,nishio2019client,bagdasaryan2018backdoor}, while the remaining one is a real-world input corpus collected from our IMA, named as \dataset.
\dataset{} contains texts input from the devices covered in the state traces in $\S$\ref{subsec:datasets}.\footnote{Due to privacy concerns, we do not include \dataset{} in our GitHub repository.} 
Each dataset can be used for an FL task. Specifically, Femnist and Celeba are for image classification tasks, while Reddit and \dataset{} are for next-word prediction tasks.
For Femnist and Celeba, we use CNN models, and for Reddit and \dataset{}, we use LSTM models. The four models are implemented by \textit{Leaf}~\cite{caldas2018leaf}, a popular FL benchmark.
All the datasets are non-IID datasets, i.e., the data distribution is skewed and unbalanced across devices, which is a common data distribution in FL scenarios~\cite{kairouz2019advances}.
We randomly split the data in each device into a training/testing set (80\%/20\%).

\bfsubsubsection{Ethic considerations}
All the data are collected with explicit agreements with users on user-term statements and a strict policy in data collection, transmission, and storage.
The IMA users are given an explicit option to opt-out of having their data collected.
In addition, we take very careful steps to protect user privacy and preserve the ethics of research.
First, our work is approved by the Research Ethical Committee of the institutes that the authors are currently affiliated with.
Second, the users’ identifies are all completely anonymized during the study.
Third, the data are stored and processed on a private, HIPPA-compliant cloud server, with strict access authorized by the company that develops the IMA. The whole process is compliant with the privacy policy of the company.

\subsection{The Simulation Platform}
\label{subsec:fl-runtime}
\reviewer{A simulation based method is highly questionable.}

\begin{figure}
    \centering
    \includegraphics[width=\linewidth]{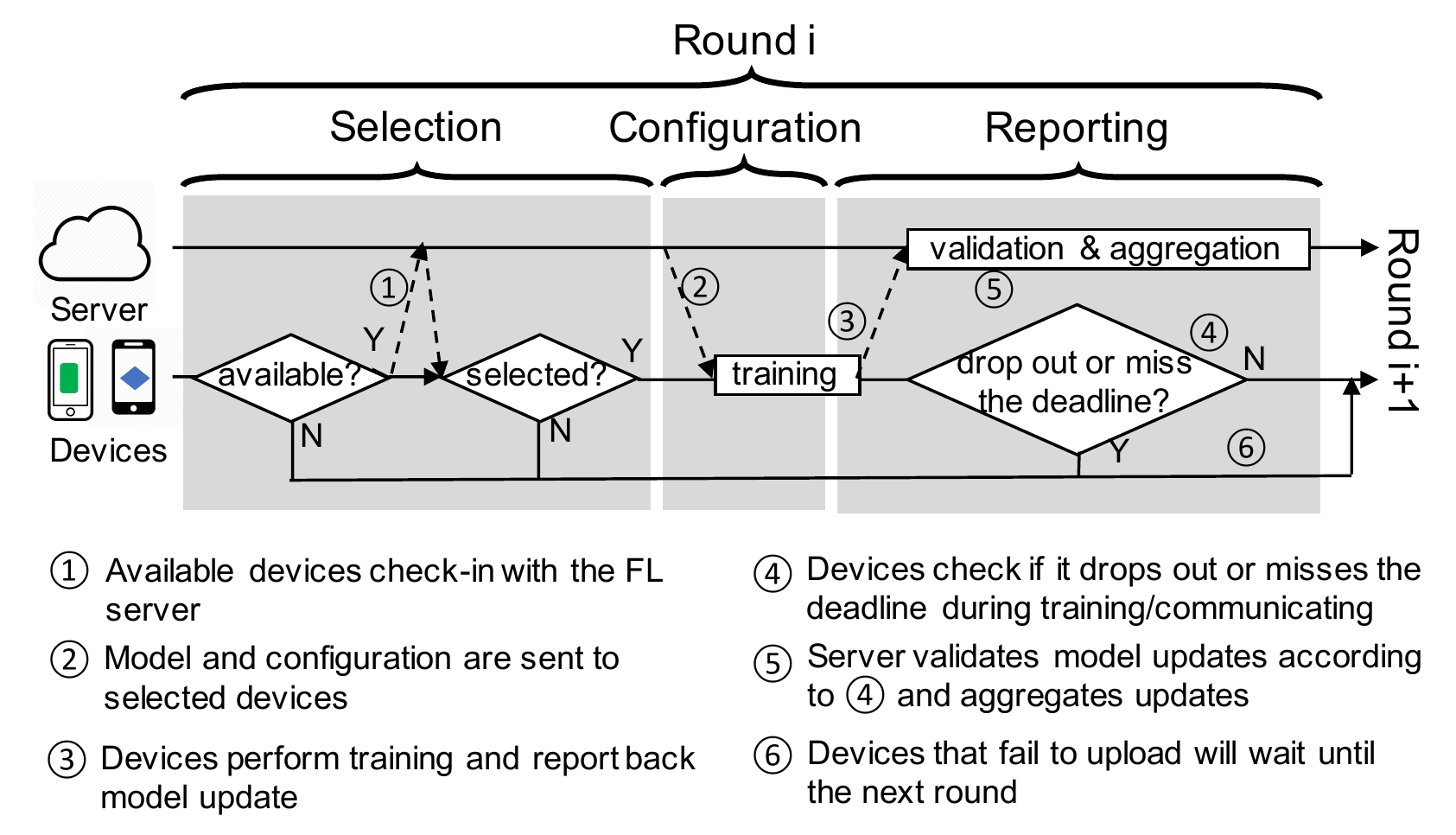}
    \caption{We build our FL simulation platform atop the standard FL protocol~\cite{bonawitz2019towards}.}
    \label{fig:fl-workflow}
\end{figure}

\noindent Our platform follows the standard FL protocol~\cite{bonawitz2019towards} and divides the simulation into three main following phases as shown in Figure \ref{fig:fl-workflow}.
We also follow the Google's report~\cite{yang2018applied} to configure the FL systems
, e.g., the time that the server waits for devices to check-in.
Given an FL task, a global ML model is trained in a synchronized way and advanced round by round.

\noindent \textbf{Selection.}
At the beginning of each round, the server waits for tens of seconds for devices to check-in.
Devices that meet the required state criteria check in to the server (\textcircled{1}).
Then the server randomly selects a subset (by default 100) of these training-available devices.

\noindent \textbf{Configuration.}
The server sends the global model and configuration to each of the selected devices (\textcircled{2}). 
The configuration is used to instruct the device to train the model.
The device starts to train the model using its local data once the transmission completed (\textcircled{3}).

\noindent \textbf{Reporting.}
The server waits for the participating devices to report updates. 
The time that the server waits is configured by the \textit{reporting deadline}.
Each device first checks its ``reporting qualification'' (\textcircled{4}), i.e., whether it has dropped out according to its states over the corresponding time period.
It also checks if it has missed the deadline according to the time needed to finish training and communication.
The preceding checking is powered by our IMA dataset described in $\S \ref{subsec:datasets}$.
The server validates updates based on the checking results and aggregates the qualified updates (\textcircled{5}).
Devices that fail to report and those that are not selected will wait until the next round (\textcircled{6}).
This reporting qualification step is what enables \ha{} FL and distinguishes our platform from existing ones.

\subsection{Experimental Settings}
\label{subsec:settings}
\noindent \textbf{Algorithms}.
We briefly introduce the algorithms explored in our study and leave more details and their hyper-parameters in $\S$\ref{sec:result}.
The algorithms can be divided into three categories:
(1) The Basic algorithm, i.e., \textit{FedAvg}, which has been deployed to real systems~\cite{bonawitz2019towards} and is widely used in FL literature~\cite{li2019fair, mohri2019agnostic, jiang2019improving, konevcny2016federatedstrategies}.
(2) Aggregation algorithms that determine how to aggregate the weights/gradients uploaded from multiple devices, including \textit{q-FedAvg}~\cite{li2019fair} and \textit{FedProx}~\cite{li2018federated}. 
(3) Compression algorithms, including \textit{Structured Updates}~\cite{konevcny2016federatedstrategies}, \textit{Gradient Dropping} (\textit{GDrop})~\cite{GDrop}, and \textit{SignSGD}~\cite{SignSGD}, which compress local models' weights/gradients to reduce the communication cost between devices and the central server. 

\noindent \textbf{Metrics}.
In our experiments, we quantify the impacts of \challengeit{} by reporting the following metrics:
(1) \textit{Convergence accuracy}, which is directly related to the performance of an algorithm.
(2) \textit{Training time/round}, which is defined as the time/rounds for the global model to converge.
Noting that the training time reported by our simulation platform is the running time after the FL system is deployed in real world instead of the time to run simulation on the pure cloud.
(3) \textit{Compression ratio}, which is defined as the fraction of the size of compressed gradients to the original size~\cite{compratio}.
(4) \textit{Variance of accuracy}, which is calculated as the standard deviation of accuracy across all the devices in the benchmark dataset. This metric indicates the cross-device fairness of an algorithm.
Table~\ref{tab:alg_metrics} summarizes the algorithms and their corresponding metrics that we measure.

\begin{table}[]
    \centering
    \scalebox{0.85}{
    \begin{tabular}{c|cccc}
        \toprule
        Algorithms & Acc. &\makecell{Training\\Time/Round} & \makecell{Compression\\Ratio} & \makecell{Var. of Acc.} \\
        \hline
        \textit{FedAvg} & $\checkmark$ & $\checkmark$ & $-$ & $-$ \\
        \hline
        \textit{Structured Updates} & $\checkmark$ & $\checkmark$ & $\checkmark$ & $-$\\
        \textit{GDrop} & $\checkmark$ & $\checkmark$ & $\checkmark$ & $-$\\
        \textit{SignSGD} & $\checkmark$ & $\checkmark$ & $\checkmark$ & $-$\\
        \hline
        \textit{q-FedAvg} & $\checkmark$ & $\checkmark$ & $-$ & $\checkmark$ \\
        \textit{FedProx} & $\checkmark$ & $\checkmark$ & $-$ & $-$ \\
        \bottomrule
    \end{tabular}
    }
    \caption{Three categories of FL algorithms we choose and their corresponding metrics we measure.}
    \label{tab:alg_metrics}
    \vspace{-15pt}
\end{table}

\noindent \textbf{Computing Environment}.
All experiments are performed on a high-performance computing cluster with Red Hat Enterprise Linux Server release 7.3 (Maipo).
The cluster has 10 GPU workers.
Each worker is equipped with 2 Intel Xeon E5-2643 V4 processor, 256G main memory, and 2 NVIDIA Tesla P100 graphics cards.
In total, the reported experiments cost more than 5,700 GPU-hours.
\section{Results}\label{sec:result}
\noindent In this section, we report the results on how \challengeit{} impacts the performance of the basic \textit{FedAvg} algorithm ($\S$\ref{subsec:basic}) and advanced FL algorithms proposed by recent FL-related studies ($\S$\ref{subsec:advance}).

\subsection{Impacts on Basic Algorithm's Performance}\label{subsec:basic}
\begin{figure*}
    \centering
    \includegraphics[width=0.95\textwidth]{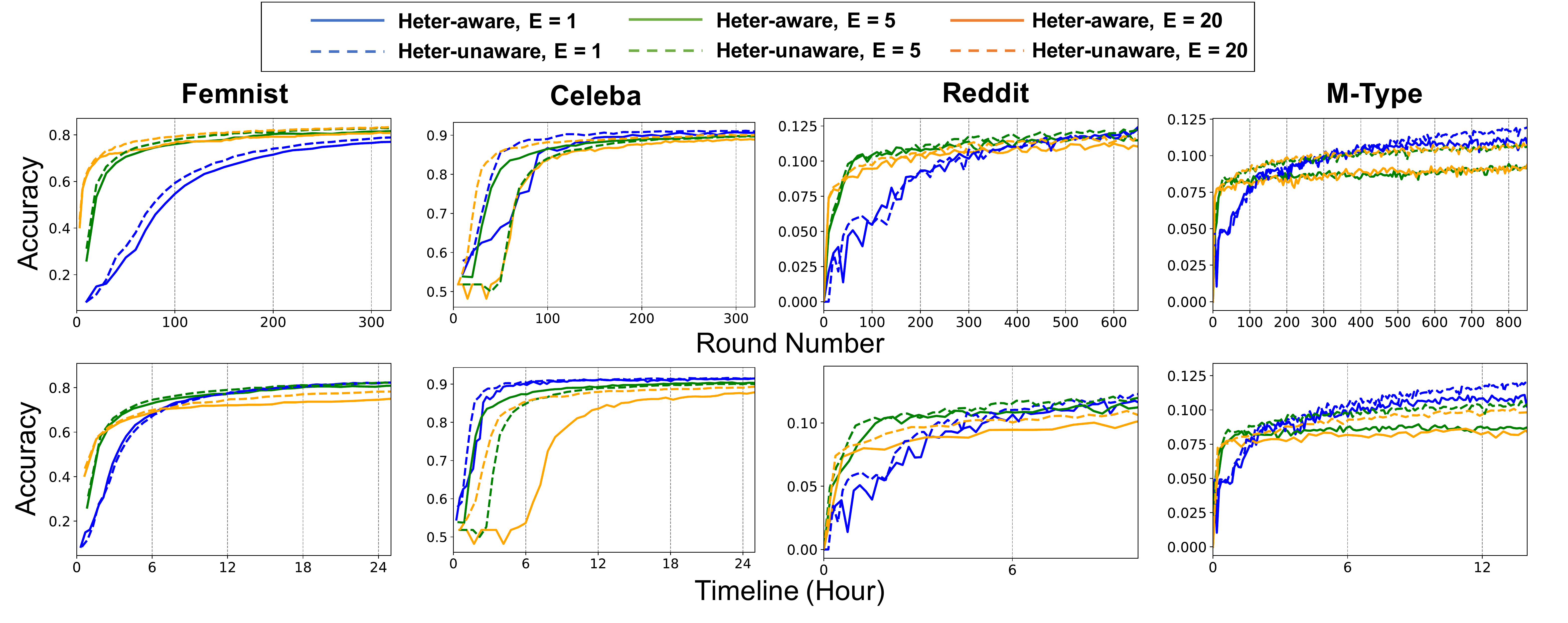}
    \caption{The testing accuracy over time, across different numbers of local training epochs (denoted as \textit{E})}.
    \label{fig:exp_1_acc}
\end{figure*}
\noindent We first measure the impacts of \challengeit{} on the performance (in terms of accuracy and training time/rounds) of the basic \textit{FedAvg} algorithm.
To obtain a more reliable result, we perform the measurement under different numbers of local training epochs, i.e., different numbers of times that the devices use their local data to update the weights of their local models (refer to $\S$\ref{sec:background}). The number of local training epoch is an important hyper-parameter of \textit{FedAvg} used to balance the communication cost between the server and the devices~\cite{mcmahan2016communication, jiang2019improving,li2018federated}. We follow previous work~\cite{mcmahan2016communication} to set this number (denoted as \textit{E}) to 1, 5, and 20. 
Also, we use the learning rate and the batch size recommended by \textit{Leaf}~\cite{caldas2018leaf} for each ML model.
Figure~\ref{fig:exp_1_acc} illustrates how accuracy changes with training time and training rounds under different numbers of local training epochs. 
We summarize our observations and insights as follows.

\noindent $\bullet$
\textbf{Heterogeneity causes non-trivial accuracy drop in FL.}
Under heterogeneity-aware settings, the accuracy drops on each dataset across various local training epoch. Specifically, the accuracy drops by an average of 2.3\%, 0.5\%, and 4\% on the existing Femnist, Celeba, and Reddit datasets, respectively. On our \dataset~dataset, the accuracy drop is more significant, with an average of 9.2\%.

\noindent $\bullet$ 
\textbf{Heterogeneity obviously slows down the training process of FL in terms of both training time and training rounds.} 
We first analyze the results in terms of training time. Under each setting of the local training epoch, the training time increases on each dataset when heterogeneity is considered. The increase ranges from 1.15$\times$ (Reddit with $E=1$) to 2.32$\times$ (Celeba with $E=20$), with an average of 1.74$\times$. In addition, we find that the training time increases more obviously when the number of local training epochs increases. When we set $E$ to 20, the training time even increases by around 12 hours on Femnist and Celeba. 
We next analyze the results in terms of training rounds. Similar to the training time, training rounds increase on each dataset when heterogeneity is considered. The increase ranges from 1.02$\times$ (\dataset{} with $E=20$) to 2.64$\times$ (Celeba with $E=20$), with an average of 1.42$\times$.

\begin{table}[]
\centering
\scalebox{0.68}{
\begin{tabular}{c|l|l|cccl}
\toprule
\textbf{Dataset} & \multicolumn{1}{c|}{\textbf{Heter.}} & \multicolumn{1}{c|}{\textbf{Algo.}} & \multicolumn{1}{c}{\textbf{Average}} & \multicolumn{1}{c}{\textbf{Worst 10\%}} & \multicolumn{1}{c}{\textbf{Best 10\%}} & \multicolumn{1}{c}{\textbf{Var. $\times 10^{-4}$ }} \\ \hline
\multirow{4}{*}{Femnist} & \multirow{2}{*}{Unaware} & \textit{FedAvg} & 82.13\% & 61.1\% & \textbf{97.2\%} & 213 \\  
 &  & \textit{q-FedAvg} & \textbf{82.66\%} & \textbf{64.7\%} & 95.1\% & \textbf{157 ($26.3\%\downarrow$)} \\  \cline{2-7}
 & \multirow{2}{*}{Aware} & \textit{FedAvg} & 81.22\% & 61.1\% & 94.9\% & 203 \\  
 &  & \textit{q-FedAvg} & \textbf{81.24\%} & \textbf{64.7\%} & \textbf{95.1\%} & \textbf{159 ($21.7\%\downarrow$)} \\ 
 
\hline
 
\multirow{4}{*}{M-Type} & \multirow{2}{*}{Unaware} & \textit{FedAvg} & \textbf{8.15\%} & 2.33\% & \textbf{13.5\%} & 19 \\  
 &  & \textit{q-FedAvg} & 7.78\% & \textbf{2.33\%} & 13.0\% & \textbf{17 ($10.5\%\downarrow$)} \\  \cline{2-7}
 & \multirow{2}{*}{Aware} & \textit{FedAvg} & 7.47\% & 2.27\% & 12.3\% & 16.2 \\  
 &  & \textit{q-FedAvg} & \textbf{7.47\%} & \textbf{2.33\%} & \textbf{12.4\%} & \textbf{15.6 ($3.7\%\downarrow$)} \\ 
 \bottomrule
\end{tabular}
}
\caption{Test accuracy for \textit{q-FedAvg} and \textit{FedAvg}. ``Var'' represents the variance of accuracy across devices.}
\label{tab:q-fedavg}
\end{table}

\subsection{Impacts on Advanced Algorithms' Performance}\label{subsec:advance}
\noindent We now measure the impacts of heterogeneity on advanced FL algorithms, e.g., model aggregation and gradient compression.

\begin{table*}[]
    \centering
    \scalebox{0.85}{
    \begin{tabular}{clrrrrrr}
         \toprule
         \makecell{Dataset} & 
         \makecell{Algo.} & 
         \makecell{Acc  (\%)\\Heter-unaware} & \makecell{Acc  (\%)\\Heter-aware} &
         \makecell{Acc Change\\(ratio)} &
         \makecell{Training time\\Heter-unaware}&
         \makecell{Training time\\Heter-aware}&
         \makecell{Compression\\Ratio}\\
         \hline
         \multirow{5}*{Femnist} & No Compression & $84.1$ ($0.0\%$) & $83.0$ ($0.0\%$) & $1.2\%\downarrow$ & 5.56 hours (1.0$\times$)& 5.96 hours (1.0$\times$)& $100\%$\\

         & Structured Updates & $\textbf{84.2}$ ($0.1\%\uparrow$) & $\textbf{83.2}$ ($0.3\%\uparrow$) & $1.1\%\downarrow$ & \textbf{5.23 hours (0.95$\times$)} & \textbf{5.56 hours (0.93$\times$)} & $6.7\%$\\ 
         
         & GDrop & $82.2$ ($2.2\%\downarrow$) & $81.5$ ($1.8\%\downarrow$) & $0.8\%\downarrow$ & 7.17 hours (1.3$\times$)& 7.98 hours (1.3$\times$)& $21.4\%\sim28.2\%$ \\ 
         
         & SignSGD & $79.0$ ($6.1\%\downarrow$) & $76.3$ ($8.1\%\downarrow$) & $3.4\%\downarrow$ & 7.62 hours (1.4$\times$)& 20.5 hours (3.4$\times$) & $\textbf{3.1\%}$ \\
         
         
         &&&&&&& \\
         
         \multirow{5}*{M-Type} & No Compression & $9.86$ ($0.0\%$) & $9.28\%$ ($0.0\%$) & $5.9\%\downarrow$ & 0.54 hours (1.0$\times$)& 1.23 hours (1.0$\times$)& $100\%$\\ 
         
         
         & Structured Updates & $9.93$ ($0.6\%\uparrow$) & $9.08$ ($2.2\%\downarrow$) & $8.6\%\downarrow$ & \textbf{0.53 hours (0.98$\times$)} & \textbf{1.59 hours (1.3$\times$)} & $39.4\%$\\ 
         
         & GDrop & $8.09$ ($18.0\%\downarrow$) & $8.27$ ($10.9\%\downarrow$) & $2.2\%\uparrow$ & 5.34 hours (10.0$\times$)& 4.29 hours (3.5$\times$)& $\textbf{0.1}\%\sim\textbf{2.1}\%$ \\ 
         
         & SignSGD & $\textbf{10.4}$ ($6.0\%\uparrow$) & $\textbf{9.55}$ ($2.9\%\uparrow$) & $8.5\%\downarrow$ & 1.45 hours (2.7$\times$)& 3.93 hours (3.2$\times$) & $3.1\%$ \\
         \bottomrule
    \end{tabular} 
    }
    \caption{
    The performance of different gradients compression algorithms.
    Numbers in the brackets indicate the accuracy change compared to the ``No Compression'' baseline.
    ``Acc. Change'' refers to the accuracy change introduced by heterogeneity.
    The compression ratio is the fraction of the size of compressed gradients to the original size.
    }
    \label{tab:grad_comp}
    \vspace{-15pt}
\end{table*}

\bfsubsubsection{Aggregation Algorithms}
\label{subsec:aggr}
The aggregation algorithm is a key component in FL that determines how to aggregate the weights or gradients uploaded from multiple devices.
Besides \textit{FedAvg}, various aggregation algorithms are proposed to improve efficiency~\cite{li2018federated,nishio2019client,niu2019secure}, ensure fairness~\cite{li2019fair}, preserve privacy~\cite{bonawitz2017practical,niu2019secure}, etc. 
To study how \challengeit~affects the performance of aggregation algorithms, we focus on two representative ones: \textit{q-FedAvg}~\cite{li2019fair} and \textit{FedProx}~\cite{li2018federated}, both of which are open-sourced.
\textit{q-FedAvg} is proposed to address the fairness issues in FL. It minimizes an aggregated reweighted loss so that the devices with higher loss are given higher relative weights.
\textit{FedProx} is proposed to tackle with \hh~in FL.
Compared to \textit{FedAvg}, \textit{FedProx} allows devices to perform various amounts of training work based on their available system resources, while \textit{FedAvg} simply drops the stragglers that fail to upload the model updates.
\textit{FedProx} also adds a proximal term to the local optimization objective (loss function) to limit the impact of variable local updates.

\begin{figure}
    \centering
    \includegraphics[width=\linewidth]{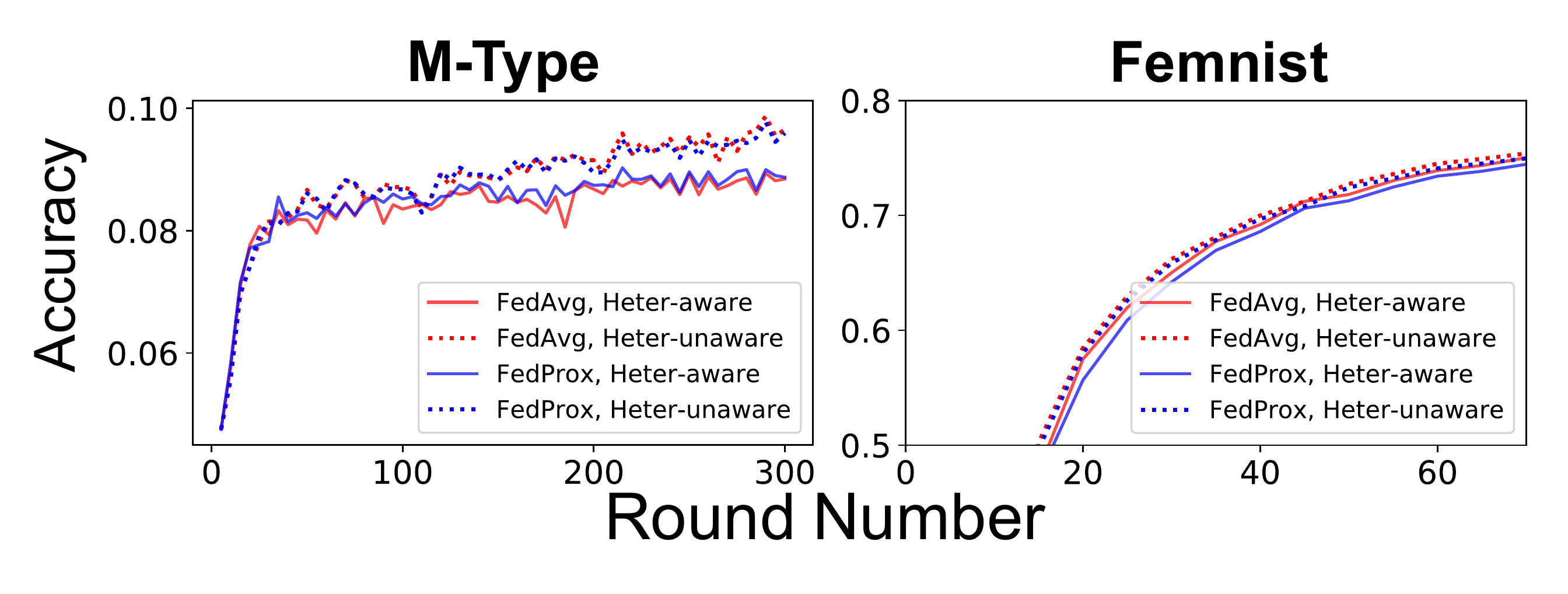}
    \vspace{-20pt}
    \caption{The training performance of \textit{FedProx} and \textit{FedAvg} with and without heterogeneity.}
    \label{fig:fedprox}
    \vspace{-10pt}
\end{figure}

We use \textit{FedAvg} as the baseline for comparison.
Due to the different optimization goals of \textit{q-FedAvg} and \textit{FedProx}, we make the comparison separately. For \textit{q-FedAvg}, the results are shown in Table \ref{tab:q-fedavg}, which illustrates the the same metrics as evaluated by \textit{q-FedAvg} (variance of accuracy, worst 10\% accuracy, i.e., 10\% quantile of accuracy across devices, and best 10\% accuracy, i.e., 90\% quantile of accuracy across devices). For \textit{FedProx}, the results are shown in Figure \ref{fig:fedprox}, which presents the accuracy changes by round.
Due to space limit, we show only the results on two datasets, i.e., one dataset using the CNN model (Femnist) and another dataset using the LSTM model (\dataset{}). Our observations are as follows.

\noindent $\bullet$ \textbf{\textit{q-FedAvg} that is supposed to address fairness issues is less effective in ensuring fairness under \ha{} settings.}
According to Table \ref{tab:q-fedavg}, under \hua{} settings, the worst 10\% accuracy of \textit{q-FedAvg} is higher than that of \textit{FedAvg} and \textit{q-FedAvg} also obtains lower variance of accuracy on both datasets.
However, under \ha{} settings, the variance reduction decreases from 26.3\% to 21.7\% on Femnist and from 10.5\% to 3.7\% on \dataset, respectively.
It is probably because \textit{q-FedAvg} cannot tackle the bias in device selection introduced by \bh{} (see details in $\S\ref{subsec:bias}$), which makes \textit{q-FedAvg} less effective in ensuring fairness.

\noindent $\bullet$ \textbf{\textit{FedProx} is less effective in improving the training process with heterogeneity considered.} 
According to Figure \ref{fig:fedprox}, on \dataset, \textit{FedProx} only slightly outperforms \textit{FedAvg}, and the heterogeneity causes an accuracy drop of 7.5\%.
On Femnist, \textit{FedProx} achieves the same performance as \textit{FedAvg} under \hua{} settings and slightly underperforms \textit{FedAvg} under \ha{} settings. The \challengeit{} causes an accuracy drop of 1.2\%.
Note that \textit{FedProx} incorporates \hh~into its design while leaving \bh~unsolved.
We manually check the involved devices and find that only 51.3\% devices have attended the training when the model reaches the target accuracy.
As a result, the model may have been dominated by these active devices and perform badly on other devices.


\bfsubsubsection{Gradient Compression Algorithms}
\label{subsec:grad_comp}
The cost of device-server communication is often reported as a major bottleneck in FL~\cite{kairouz2019advances}, so we first investigate gradient compression algorithms that are extensively studied to reduce the communication cost.
Specifically, we focus on three well-adopted gradient compression algorithms: \textit{Structured Updates}~\cite{konevcny2016federatedstrategies}, \textit{Gradient Dropping} (\textit{GDrop})~\cite{GDrop}, and \textit{SignSGD}~\cite{SignSGD}.
For each of them, we tune the hyper-parameters to achieve the highest accuracy through massive experiments. As a result,
for \textit{Structured Updates}, we set the max rank of the decomposited matrix to 100;
for \textit{GDrop}, we set the weights dropout threshold to 0.005;
for \textit{SignSGD}, we set the learning rate to 0.001, the momentum constant to 0, and the weight decay to 0. We use \textit{FedAvg} with no compression as the baseline for comparison.
Besides accuracy and training time/rounds, we also use compression ratio (described in $\S$\ref{subsec:settings}) as the measurement metrics of these algorithms. We present the metric values of the three compression algorithms as well as the baseline under heterogeneity-unaware and heterogeneity-aware settings in Table~\ref{tab:grad_comp}. 
Similar to $\S\ref{subsec:aggr}$, we report only the results on Femnist and \dataset.
We summarize our findings as follows.

\noindent $\bullet$ \textbf{Heterogeneity introduces a similar accuracy drop to compression algorithms as it does to the basic algorithm.}
We measure the accuracy change introduced by \challengeit{}~(noted as \textit{Acc. Change} in Table \ref{tab:grad_comp}).
We observe that the introduced accuracy degradation ($3.1\%$ on average) is similar to the one ($3.2\%$ on average) that we observe in $\S\ref{subsec:basic}$.
On average, the accuracy drops by 1.7\% on Femnist and 5.3\% on \dataset.
It is reasonable because \challengeit{} will not affect the compressed gradients.


\noindent $\bullet$ \textbf{Gradient compression algorithms can hardly speed up the model convergence under \ha{} settings.}
Although all these algorithms compress the gradients and reduce the communication cost significantly (the compression ratio ranges from 0.1\% to 39.4\%), the training time is seldom shortened (only \textit{Structured Updates} shortens the convergence time to 0.93$\times$ at most) and lengthened in most cases.
For example, on \dataset{} under \ha~environment, the training time is lengthened by 1.3$\times$ to 2.5$\times$ for all compression algorithms.
The training time has not been shortened for two reasons.
First, we find that communication accounts for only a small portion of the total learning time compared to on-device training.
Most devices can finish the downloading and uploading in less than 30 seconds for a model around 50M while spending more time (1-5 minutes with 5 epochs) on training.
Second, the accuracy increases slowly when the gradients are compressed and the \challengeit{} is introduced (refer to $\S\ref{subsec:basic}$), thus taking more rounds to reach the target accuracy.



\begin{figure*}
    \centering
    \includegraphics[width=0.95\linewidth]{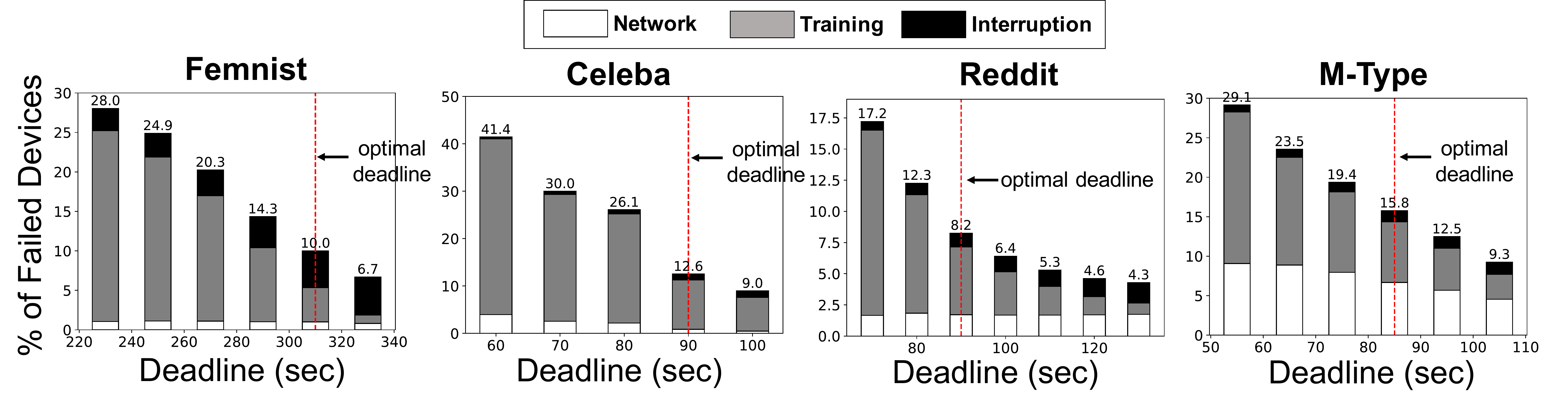}
    \caption{The prevalence of different failure reasons. The optimal deadline (red line) refers to the one that achieves the shortest training time.}
    \label{fig:failure}
\end{figure*}
\begin{figure}
    \centering
    \includegraphics[width=0.98\linewidth]{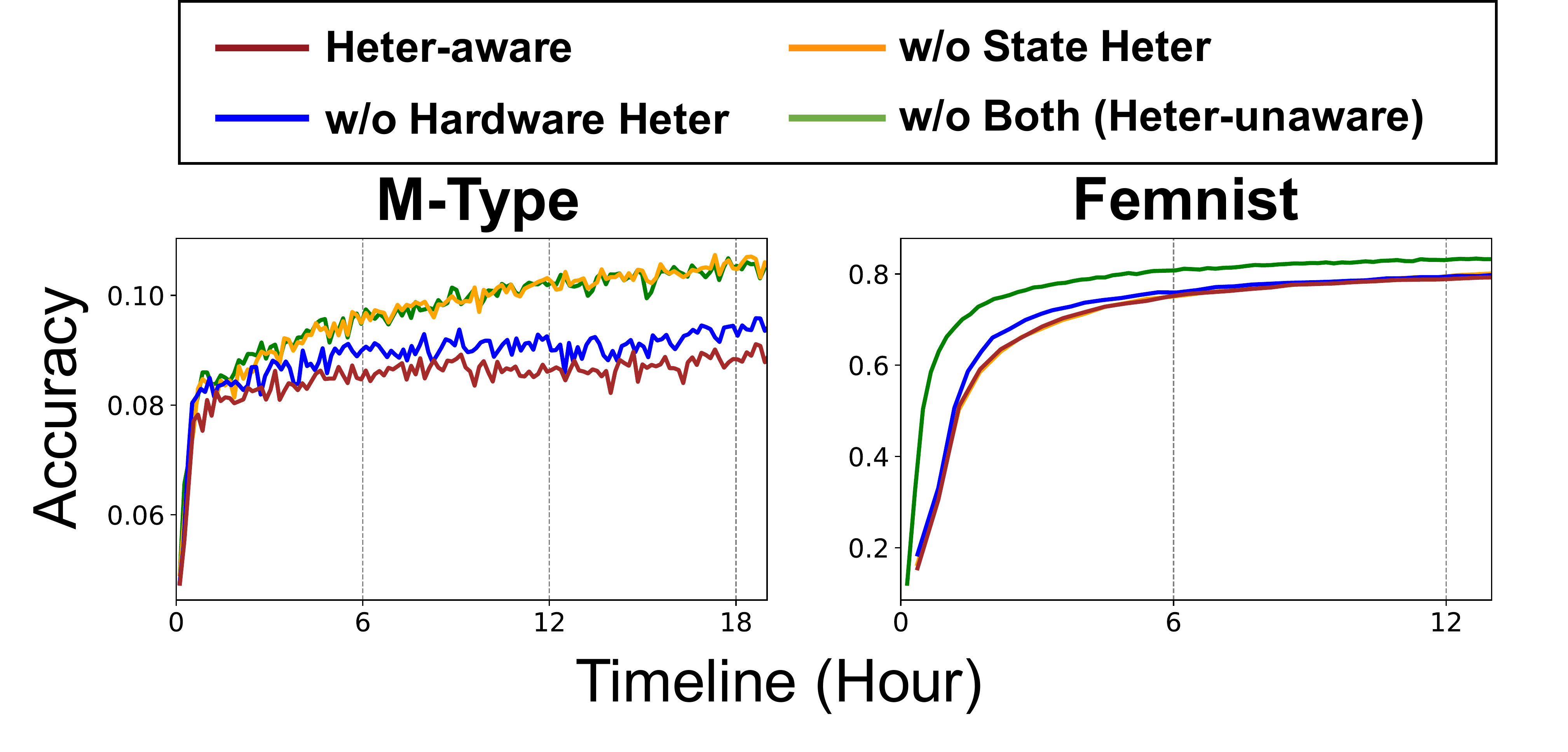}
    \vspace{-10pt}
    \caption{A breakdown of the impacts of different types of heterogeneity. \Bh{} causes more performance degradation than \hh{}. ``Heter'' is short for heterogeneity.}
    \label{fig:hete_acc}
\end{figure}
\section{Analysis of Impact Factors} \label{sec:cause}
\noindent Given the non-trivial negative impacts of heterogeneity shown in the previous section, we dive deeper to analyze the main factors of these impacts. In this section, we focus on \textit{FedAvg}, considering its wide usage in practical applications. Specifically, we first break down heterogeneity into two types, i.e., \bh{} and \hh{}, to analyze their individual impacts ($\S\ref{subsec:breakdown}$). Then we report two phenomena that are particularly obvious under \ha{} settings according to our experiments: (1) selected devices can fail to upload their model updates for several reasons, which we call \textit{device failure} ($\S\ref{subsec:failure}$); (2) devices that succeed in uploading still have biased contribution to the global model, which we call \textit{participant bias} ($\S\ref{subsec:bias}$).

\subsection{Breakdown of Heterogeneity}
\label{subsec:breakdown}

\noindent The preceding results indicate the joint impacts from two types of \challengeit.
To analyze their individual impact, we disable the \hh{}, i.e., all the devices have the same computational and communication capacity (noted as ``w/o hardware heter'').
Similarly, we disable the \bh{}, i.e., devices are always available at any time and will not drop out (noted as ``w/o state heter'').
We show the accuracy changes with the training time in Figure \ref{fig:hete_acc}.

\noindent $\bullet$ \textbf{Both \bh{} and \hh{} slow down the model convergence.} According to Figure \ref{fig:hete_acc}, \bh{} leads to comparable increase of training time to \hh{}, i.e., 1.72$\times$ vs. 1.26$\times$ on \dataset{} and 2.34$\times$ vs. 2.62$\times$ on Femnist. It is reasonable because both drop-out (introduced by \bh{}) and low-end devices (introduced by \hh{}) affect the training time.

\noindent $\bullet$ \textbf{\Bh{} is more influential than \hh{} on the model accuracy.}
As shown in Figure \ref{fig:hete_acc}, \bh~leads to a more significant accuracy drop than \hh{}, i.e., 9.5\% vs. 0.4\% on \dataset{} and 1.1\% vs. 0.1\% on Femnist.
Note that existing FL-related studies usually ignore \bh{} and only a small amount of work~\cite{li2018federated, nishio2019client, laguel2020device, chai2019towards} explores \hh{} (refer to $\S\ref{sec:background}$).
Our results show that \bh{} is more responsible for the model accuracy drop, which explains why \textit{FedProx} (it considers \hh{}) is less effective given both types of \challengeit{} (refer to $\S\ref{subsec:aggr}$).

\subsection{Device Failure}\label{subsec:failure}
\noindent Device failure refers to the phenomenon that a selected device misses the deadline to upload the model updates in a round.
It can slow down the model convergence and cause a waste of valuable device resources (computations, energy, etc.).
However, device failure is seldom studied in prior work, probably because it is directly related to the FL heterogeneity.

Heuristically, we categorize device failure to three possible causes:
(1) \textbf{\textit{Network failure}} is detected if the device takes excessively long time (default: 3$\times$ the average) to communicate with the server due to a slow or unreliable network connection.
(2) \textbf{\textit{Interruption failure}} is detected if the device fails to upload the model updates due to the user interruption, e.g., the device is uncharged during training.
(3) \textbf{\textit{Training failure}} refers to the case when the device takes too much time on training.

\begin{figure}
    \centering
    \includegraphics[width=\linewidth]{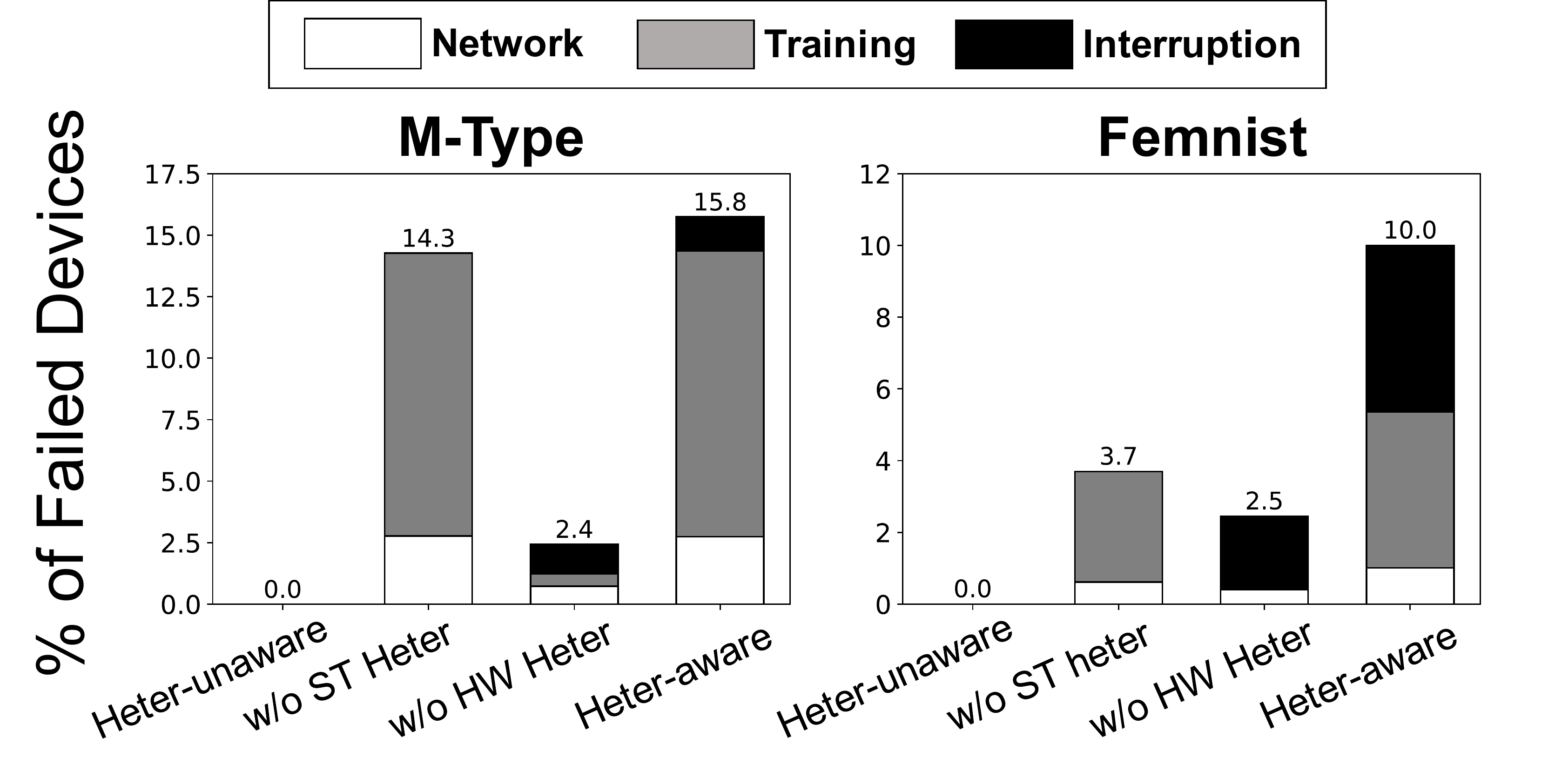}
    \caption{Different kinds of heterogeneity's influence on device failure.}
    \label{fig:hete_failure}
    \vspace{-10pt}
\end{figure}

To understand device failure, we zoom into the previous experiments under varied round deadlines. We vary the deadline because we find that the proportion of failed devices is greatly affected by it.
Similar to $\S$\ref{subsec:breakdown}, we will also check \hh's and \bh's influence on device failure.
The key questions we want to answer here are: (1) how often the devices may fail and what the corresponding reasons for the failure are;
(2) and which type of heterogeneity is the major factor.
The results are illustrated in Figures~\ref{fig:failure} and \ref{fig:hete_failure}, from which we make the following key observations.

\noindent $\bullet$ \textbf{Heterogeneity introduces non-trivial device failure even when an optimal deadline setting is given.}
The overall proportion of the failed devices reaches 11.6\% on average, with an optimal deadline setting that achieves the shortest training time.
A tight deadline increases the failure proportion because devices receive less time to finish their training tasks. We look into three types of failure and find that:
(1) Network failure accounts for a small fraction of device failure (typically less than 5\%) and it is more stable than other types of failure.
(2) Interruption failure is affected by the deadline but in a moderate way. 
We further break down the interruption failure into three sub-categories corresponding to three restrictions on training~\cite{bonawitz2019towards}. Specifically, results show that the training process is interrupted by user interaction, battery charged-off, and network changes with a probability of 46.06\%, 36.96\%, and 17.78\% respectively.
(3) Training failure is heavily affected by the deadline.
This type of failure accounts for the majority of the device failure when the deadline is set too tight.
Even with the optimal deadline setting, this type of failure still occurs because we observe that some low-end devices with too many local data sometimes fail to meet the deadline.

\noindent $\bullet$ \textbf{\Hh{} leads to more device failure than \bh.}
According to Figure \ref{fig:hete_failure}, \hh{} is more responsible for the device failure.
For example, on \dataset, \hh~causes 14\% failed devices on average while \bh~causes only 2.5\%.
It is probably because when \hh{} is considered, there are low-end devices that suffer longer training time.

\subsection{Participant  Bias}\label{subsec:bias}
\noindent Participant bias refers to the phenomenon that devices do not participate in FL with the same probability.
It can lead to different contributions to the global model, thus making some devices under-represented.
Due to \bh{}, devices frequently used by users are less likely to check in.
Due to \hh{}, low-end devices are less likely to upload their updates to the central server.

\begin{figure}
    \centering
    \includegraphics[width=\linewidth]{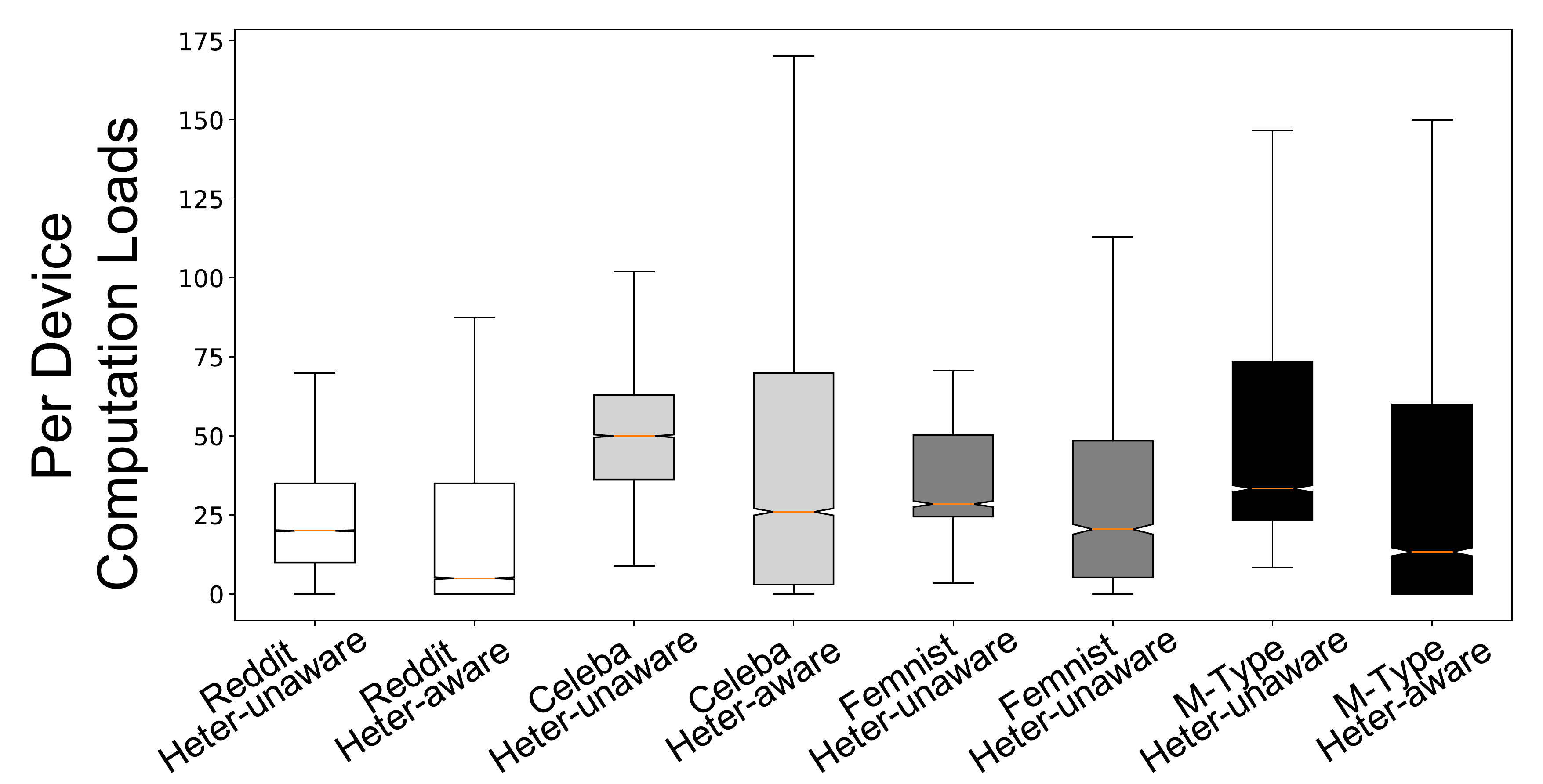}
    \caption{The distribution of computations across devices during FL training.}
    \label{fig:comp}
    \vspace{-10pt}
\end{figure}

To measure the participant bias introduced by \challengeit, we run the same FL tasks in $\S$\ref{subsec:basic}.
We take the amount of computation to reflect the participation degree of different devices. 
Since it is difficult to compare the computation of different models directly, we divide them by the amount of computation for a training epoch (noted as computation loads).
Figure \ref{fig:comp} illustrates the distribution of computation loads across devices when the global model reaches the target accuracy.
Similar to $\S$\ref{subsec:breakdown}, we also break down to explore the impacts of different types of heterogeneity.
We summary our findings as follows.

\begin{figure}
    \centering
    \includegraphics[width=\linewidth]{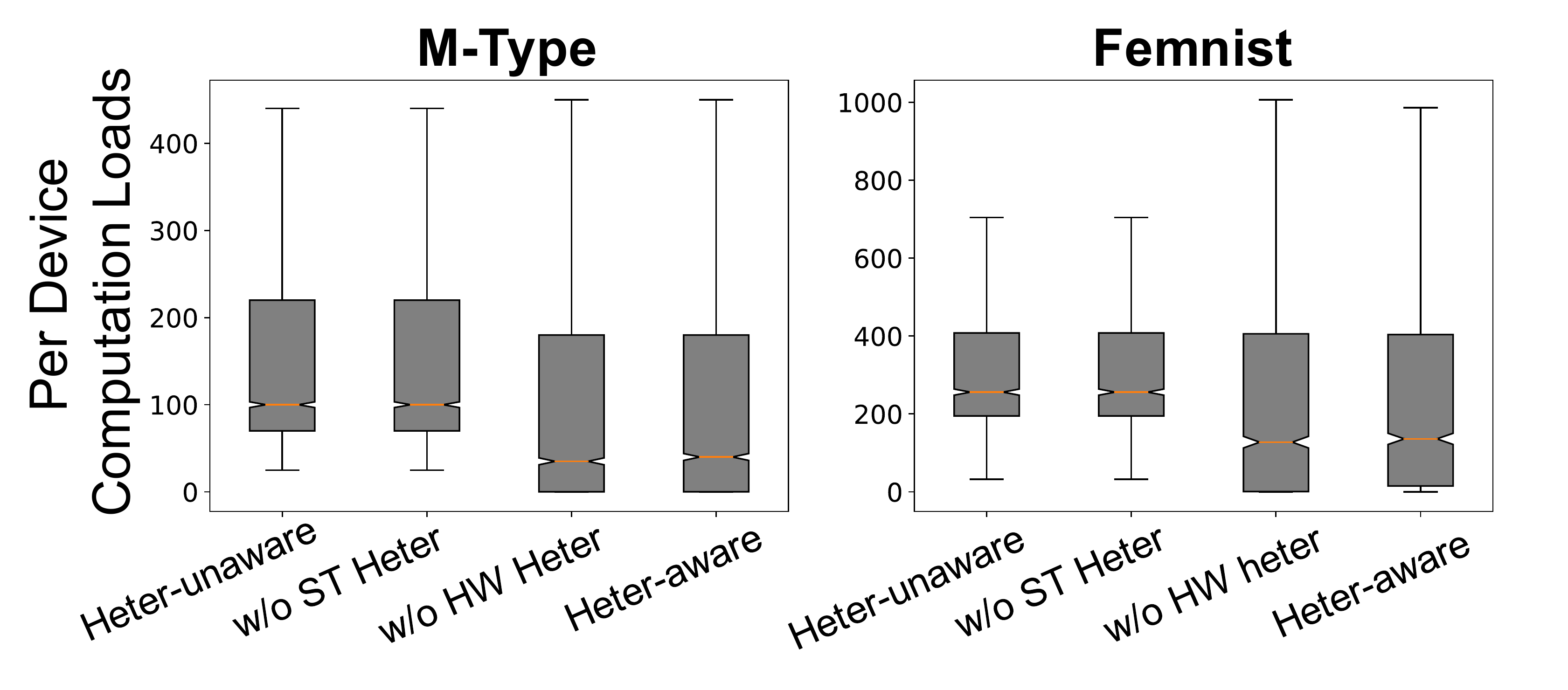}
    \caption{A breakdown of the impacts of different types of Heterogeneity on participant bias.}
    \label{fig:hete_comp}
\end{figure}

\noindent $\bullet$
\textbf{The computation loads get more uneven under \ha{} settings.} 
The variance is increased by 2.4$\times$ (Reddit) to 10.7$\times$ (Femnist).
Compared to \hua{} environment where every device participates with an equal probability, in the \ha{} environment, the computation loads have a trend of polarization.
On Celeba, the maximum computation load increases by 1.17$\times$.

\noindent $\bullet$
\textbf{The number of inactive devices increases significantly under \ha{} settings.}
The median computation load drops by 28\% (Femnist) to 75\% (Reddit), indicating that more inactive devices appear.
Compared to the \hua{} environment where top 30\% of the devices contribute 54\% of the total computation, in the \ha{} environment, top 30\% of the devices contribute 81\% of the total computation, putting the inactive devices at a disadvantage.

\begin{wrapfigure}{r}{4cm}
\centering
\vspace{-15pt}
\includegraphics[width=\linewidth]{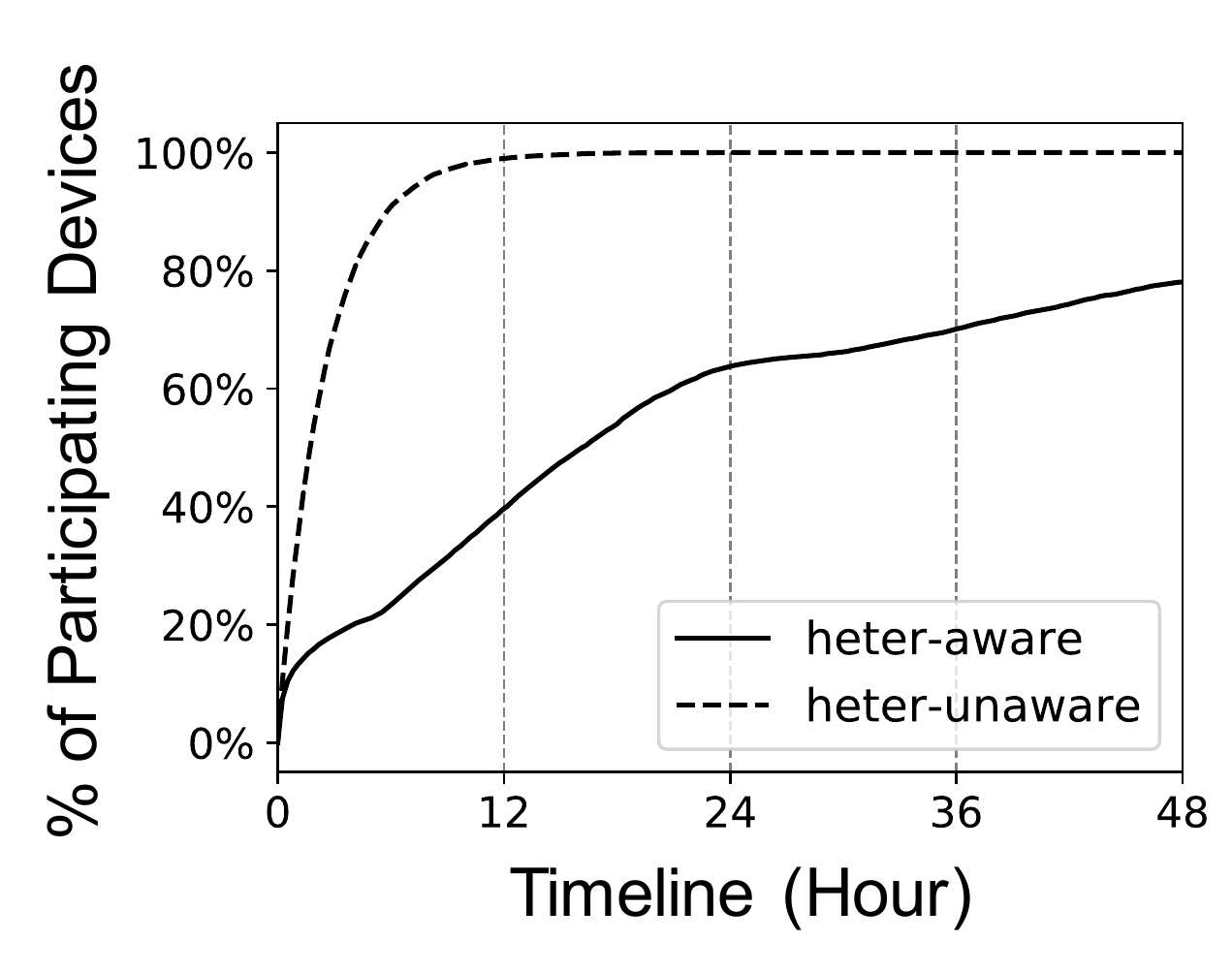}
\vspace{-15pt}
\caption{Percentage of participating devices over time.}
\label{fig:attend}
\end{wrapfigure}

\noindent $\bullet$
\textbf{Up to 30\% devices have not participated in FL process when the global model reaches the target accuracy under \ha{} settings.}
To investigate the reasons for these inactive devices, we inspect the percentage of participating devices over time and demonstrate the result in Figure \ref{fig:attend}.
We find that when the model reaches the target accuracy (6-24 hours in our experiments), more than 30\% devices have not participated.
In the \hua{} environment, the participating devices accumulate quickly and soon cover the total population in 12 hours. 
While in \ha{} environment, the accumulation speed gets much slower and it takes much longer time to converge (more than 48 hours).

\noindent $\bullet$
\textbf{\Bh{} is more responsible for participant bias.}
As shown in Figure \ref{fig:hete_comp}, \bh{} is the main reason for computation bias.
It causes the similar computation distribution as the one in \ha{} environment.
It is probably because \bh{} introduces bias in device selection, i.e., although the server selects devices randomly, the available devices that can be selected highly depend on if the device can meet the state criteria (refer to $\S\ref{subsec:datasets}$).


\section{Implications}\label{sec:impli}
\noindent In this section, we discuss actionable implications for FL algorithm designers and FL system providers based on our above findings.

\subsection{For FL Algorithm Designers}
\noindent \textbf{Taking \challengeit{} into consideration.} As demonstrated in our study, \challengeit{} introduces non-trivial accuracy drop and training slowdown in FL, as well as affects the effectiveness of some proposed methods.
These findings encourage researchers to consider \challengeit{}, especially \bh{}, when they practice on FL.
On the one hand, when designing approaches or algorithms, researchers should consider circumstances that are common in \ha{} environment but do not exist in \hua{} environment.
For example, when designing a device selection approach, researchers should be aware that some devices can be unavailable at a given time and the server cannot select as it wants.
When designing an aggregation algorithm, researchers should guarantee that the algorithm still works given inevitable device failure.
On the other hand, when evaluating FL algorithms, researchers should add necessary \challengeit{} settings in the experiments according to the targeted scenario.
For example, additional system overhead of the algorithm may further widen the gap in training time between different devices, which should be considered during the evaluation.

\textbf{Reducing device failure by a ``proactive alerting'' technique.}
In $\S\ref{subsec:failure}$, we find that around 10\% of devices fail to upload their model updates under typical settings.
The reasons include excessive training time, unstable network, and device drop-out caused by state changes. 
Existing efforts have explored dynamic deadline \cite{lismartpc} and tolerating partial work \cite{li2018federated} to handle the device failure.
However, these algorithms are inadequate to handle the failure caused by unstable network and drop-out because they are highly dependent on the device's states.
One may explore a ``proactive alerting'' technique by predicting the device's future states and network condition based on historical data.
The server should assign a low priority to the devices that are likely to drop out.
In this way, the overall device failure can be reduced and more updates can be aggregated thus saving the hardware resource and accelerating learning process.

\textbf{Resolving bias in device selections.}
In $\S\ref{subsec:bias}$, we find that the global model is dominated by some active devices (top 30\% of devices can contribute 81\% of the total computation).
The reason is that, due to \bh{}, devices do not participate in the learning process with the same probability even when they are randomly selected, and some (more than 30\% in our experiments) have never participated when the model reaches a local optimum.
To alleviate the bias in device selection, a naive approach is to set a participation time window (e.g., one day) and omit the devices that have participated in this window.
The ``fairness'' is guaranteed, but this may remarkably increase the training time of an FL task, and the length of the time window should be carefully tuned.
What is more, adjusting the local objective (loss function) or re-weighting updates can be possible alternatives.

\subsection{For FL System Providers}
\noindent \textbf{Building \ha{} platforms.}
Our results show that a \ha{} platform is necessary for developers to precisely understand how their model shall perform in real-world settings.
However, existing platforms \cite{TFF,ryffel2018generic,caldas2018leaf,PFL,he2020fedml} fail to incorporate \challengeit{} into their design.
Our work provides a reference implementation and can be easily integrated into these FL platforms.
We also encourage system providers to collect their own data that fit different scenarios to further help the FL community.

\textbf{Optimizing on-device training time, instead of optimizing compression in unmetered (e.g., WiFi) networks.}
In $\S\ref{subsec:grad_comp}$, we find that gradient compression algorithms can hardly speed up model convergence.
The time spent on communication is relatively small in the WiFi environment, compared to the time spent in training.
As a result, an orthogonal way to accelerate FL is to optimize the on-device training time.
Possible solutions include neural architecture search (NAS) \cite{JMLR:v20:18-598, wistuba2019survey} and using hardware AI accelerators like mobile GPU and digital signal processor (DSP). 

\section{Discussion}\label{sec:discuss}
\noindent We next discuss open problems along with generalizability of our study.

\begin{figure}[htbp]
    \centering
    \begin{minipage}[t]{0.47\linewidth}
        \includegraphics[width=\textwidth]{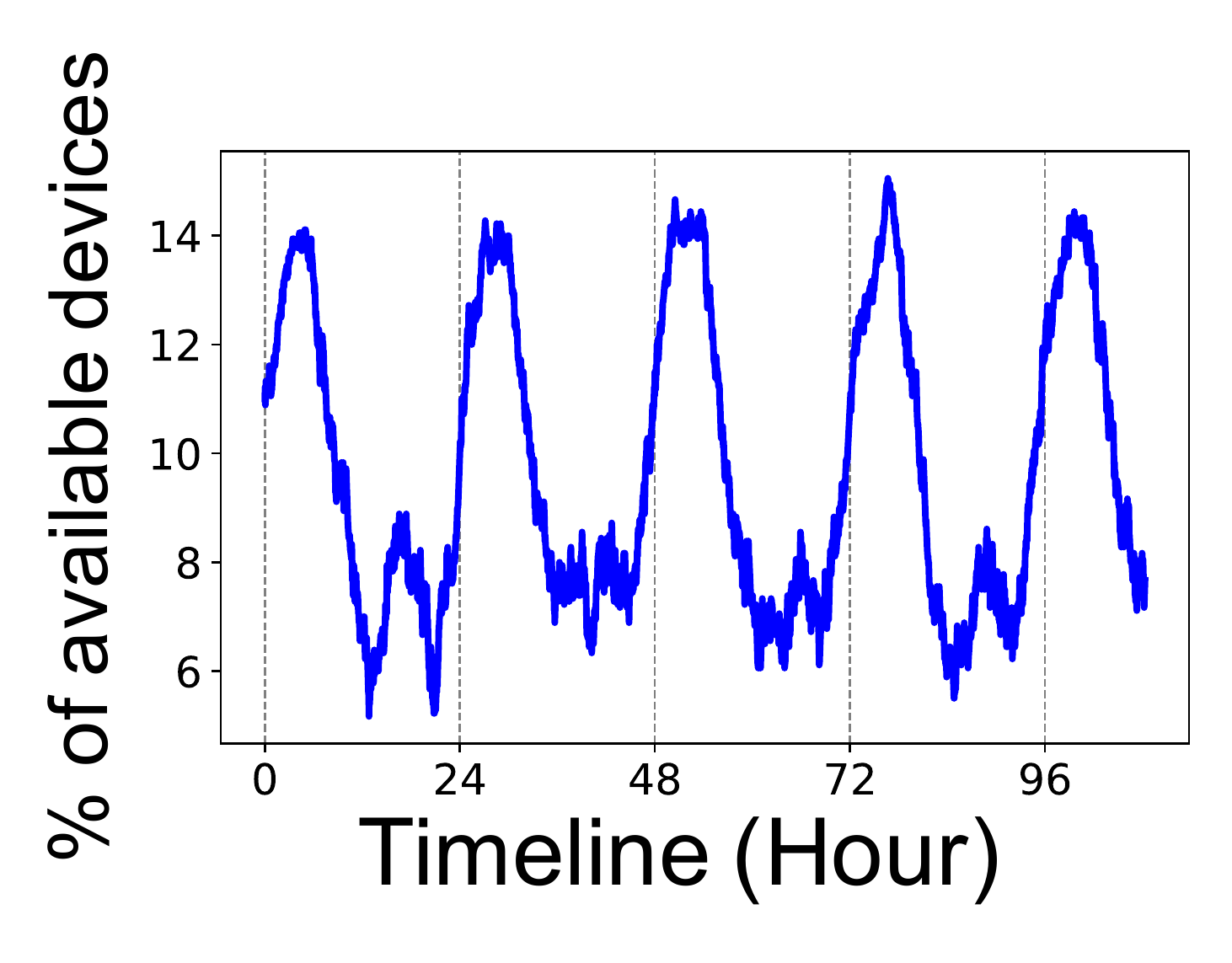}
        \vspace{-20pt}
        \caption{Percentage of available devices over time.}
        \label{fig:online_num}
    \end{minipage}
    \hspace{0.03\linewidth}
    \begin{minipage}[t]{0.47\linewidth}
        \includegraphics[width=\textwidth]{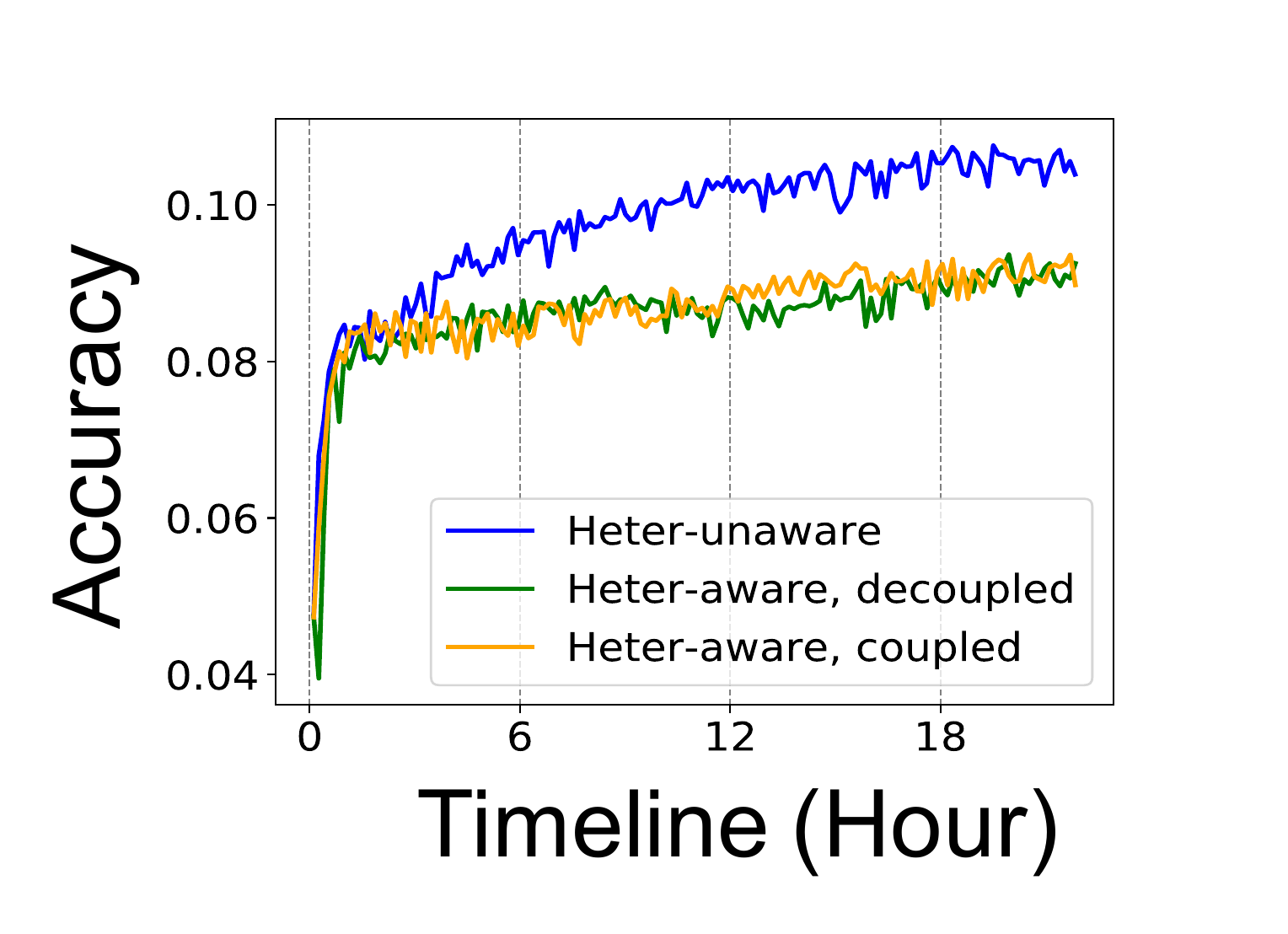}
        \vspace{-20pt}
        \caption{Decoupling is verified on \dataset.}
        \label{fig:validity}
    \end{minipage}
\end{figure}

\textbf{Bias of our IMA dataset.}
The device state traces ($\S$\ref{subsec:datasets}) are collected from our IMA (app-specific) whose users mainly reside in Southeast Asia and Latin America (geo-specific).
The traces may not be fully representative to other FL scenarios.
However, we believe that our findings are still faithful because
(1) FL task is always app-specific and improving IMA experience is a key scenario of FL~\cite{yang2018applied,bonawitz2019towards,hard2018federated};
(2) our traces are large enough to cover the general state change patterns of smartphones. What is more, the patterns are consistent with prior work~\cite{yang2018applied} as aforementioned.
Furthermore, new user traces can be seamlessly plugged into our platform where researchers can reproduce all experiments mentioned in this paper.

\textbf{Consistency with results reported by real-world FL systems.}
Similar to all existing FL platforms \cite{TFF,ryffel2018generic,caldas2018leaf,PFL,he2020fedml}, our platform ($\S\ref{subsec:fl-runtime}$) performs FL tasks in a simulation way.
We carefully design the platform to simulate the real-world FL systems by considering the \challengeit. However, we acknowledge that a gap may still exist for unexpected FL glitches, e.g., software failure.
We plan to further validate our platform with real-world deployment.
Nevertheless, the observed patterns from our platform, e.g., device availability (Figure \ref{fig:online_num}) and failure proportion (Figure \ref{fig:failure}), are consistent with the results reported from a large-scale FL deployment by Google~\cite{bonawitz2019towards}.
Therefore, we believe that our findings are still valid.

\textbf{Validity of randomly assigning state traces and training data to devices.}
In practice, the heterogeneity is inherently coupled with the non-IID data distribution~\cite{kairouz2019advances}.
In this study, we decouple the \challengeit~from the data distribution, i.e., randomly assigning a state trace to each device, to generalize our traces to other benchmark datasets.
We use \dataset{} to verify this design because it shares the same user population with our traces.
According to Figure \ref{fig:validity}, the gap between the coupled case and the decoupled case is trivial compared to the gap between the \hua~and \ha~settings.
It justifies our design to decouple \challengeit{} from any third-party datasets.

\textbf{Other types of heterogeneity.}
In this paper, we focus on the impacts of hardware and state heterogeneity. In fact, there also exist other types of heterogeneity in FL. One is data heterogeneity~\cite{li2019federated,kairouz2019advances} that resides in the skewed and unbalanced local data distribution (non-IID data distribution) across devices. Data heterogeneity is one of the basic assumptions in FL and existing work~\cite{mcmahan2016communication,konevcny2016federated,chai2019towards} has conducted in-depth research on it. Since the benchmark datasets used in our experiments are all non-IID datasets, data heterogeneity is inherently considered in our study. Other types of heterogeneity~\cite{kairouz2019advances}, like heterogeneity on software or platform, are highly relevant to the implementation of an FL system and hard to generalize. We plan to leave them for future work.

\section{Conclusion}
\noindent We have collected large-scale real-world data and conducted extensive experiments to first anatomize the impacts of heterogeneity. Results show that
(1) heterogeneity causes non-trivial performance degradation in FL tasks, up to 9.2\% accuracy drop and 2.32$\times$ convergence slowdown;
(2) recent advanced FL algorithms can be compromised and rethought with heterogeneity considered;
(3) \bh{}, which is usually ignored in existing studies, is more responsible for the aforementioned performance degradation;
(4) device failure and participant bias are two potential impact factors of performance degradation.
These results suggest that heterogeneity should be taken into consideration in further research work and that optimizations to mitigate the negative impacts of heterogeneity are promising.


\section*{Acknowledgment}
\noindent This work was supported by the National Key Research and Development Program of China under the grant number 2020YFB2104100, the National Natural Science Foundation of China under grant numbers 61725201, J1924032, and 62032003, the Beijing Outstanding Young Scientist Program under the grant number BJJWZYJH 01201910001004, and the Alibaba-PKU Joint Research Program.
Mengwei Xu was supported by the NSFC under the grant number 61921003.
\balance

\bibliographystyle{ACM-Reference-Format}
\bibliography{main}

\end{document}